\documentclass[10pt,twocolumn,letterpaper]{article}

\usepackage{3dv}
\usepackage{times}
\usepackage{epsfig}
\usepackage{graphicx}
\usepackage{amsmath}
\usepackage{amssymb}

\usepackage[pagebackref=true,breaklinks=true,letterpaper=true,colorlinks,bookmarks=false]{hyperref}
\threedvfinalcopy 

\def\threedvPaperID{141} 

\ifthreedvfinal\fi
\begin{document}

\title{Unsupervised Feature Learning for Point Cloud Understanding by Contrasting and Clustering Using Graph Convolutional Neural Networks}

\author{Ling Zhang, Zhigang Zhu\\
The City College of The City University of New York\\
{\tt\small lzhang006@citymail.cuny.edu, zzhu@ccny.cuny.edu}
\and
}

\maketitle

\begin{abstract}
To alleviate the cost of collecting and annotating large-scale "3D object" point cloud data, we propose an unsupervised learning approach to learn features from an unlabeled point cloud dataset by using part contrasting and object clustering with deep graph convolutional neural networks (GCNNs). 
In the contrast learning step, all the samples in the 3D object dataset are cut into two parts and put into a "part" dataset. Then a contrast learning GCNN (ContrastNet) is trained to verify whether two randomly sampled parts from the part dataset belong to the same object. In the cluster learning step, the trained ContrastNet is applied to all the samples in the original 3D object dataset to extract features, which are used to group the samples into clusters. Then another GCNN for clustering learning (ClusterNet) is trained from the orignal 3D data to predict the cluster IDs of all the training samples. 
The contrasting learning forces the ContrastNet to learn semantic features of objects, while the ClusterNet improves the quality of learned features by being trained to discover objects that belong to the same semantic categories by using cluster IDs. 
We have conducted extensive experiments to evaluate the proposed framework on point cloud classification tasks. The proposed unsupervised learning approach obtains comparable performance to the state-of-the-art with heavier shape auto-encoding unsupervised feature extraction methods.
We have also tested the networks on object recognition using partial 3D data, by simulating occlusions and perspective views, and obtained practically useful results.
The code of this work is publicly available at: {\color{blue}https://github.com/lingzhang1/ContrastNet}.
\end{abstract}

\section{Introduction}


With ever increasing applications, point cloud data understanding with deep graph convolutional neural networks (GCNNs) has drawn extensive attention \cite{PointNet, PointNetPlusPlus, DGCNN, pointcnn}. Various networks, such as PointNet \cite{PointNet}, PointNet++ \cite{PointNetPlusPlus}, DGCNN \cite{DGCNN} and etc., and datasets such as ModelNet \cite{modelnet}, ShapeNet \cite{ShapeNet}, and SUNCG \cite{SUNCG}, have been proposed for point cloud understanding tasks. With the help of deep models and large-scale labeled datasets, significant progress has been made on point cloud understanding tasks, including classification, segmentation and detection.

\begin{figure}[!t]
\begin{center}
\includegraphics[width=0.45\textwidth]{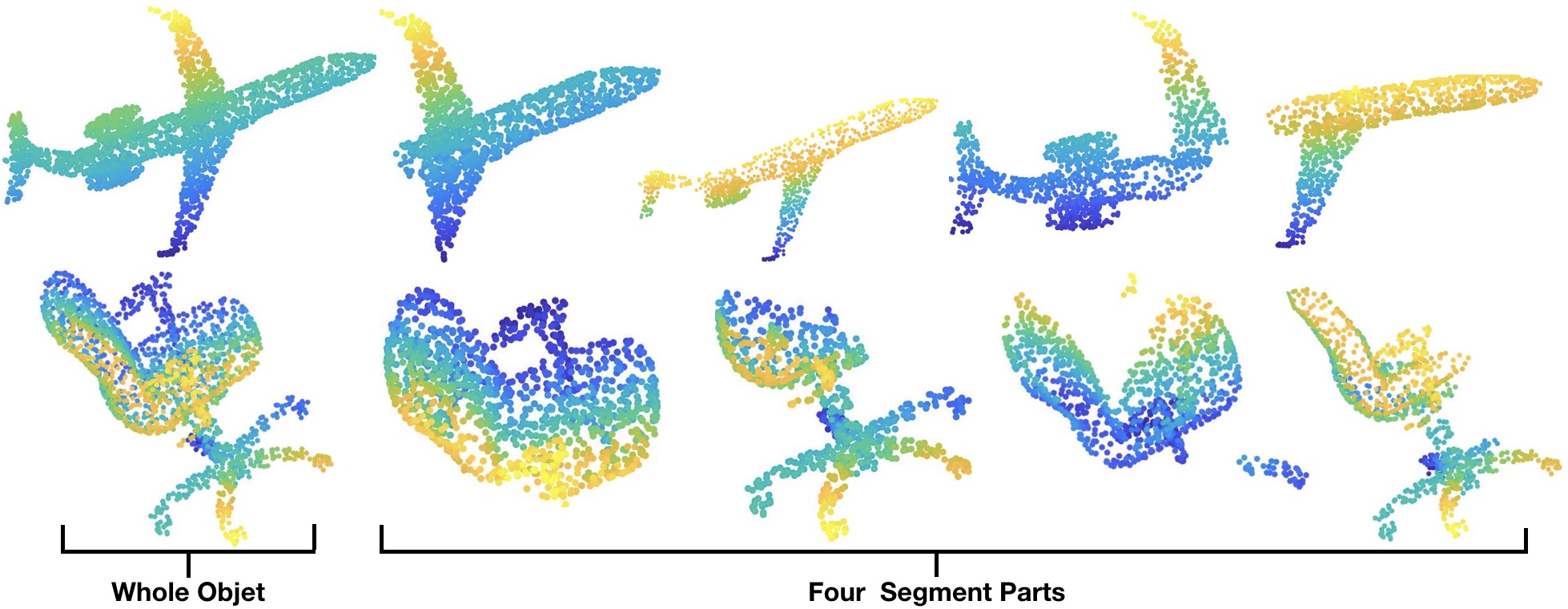}
\end{center}
\caption{Each row consists of a 3D point cloud object and its four different segments. Human can easily recognize the object and the locations of the segments in the object even for a small segment.Inspired by this observation, we propose to train GCNNs to learn features from a unlabeled dataset by recognizing whether two segments are from the same object.}
\vspace{-10pt}
\label{fig:part}
\end{figure}

GCNNs typically have millions of parameters which could easily lead to over-fitting. Large-scale annotated datasets are needed for the training of such deep networks. However, the collection and annotation of point cloud datasets are very time-consuming and expensive since pixel-level annotations are needed. With their powerful ability to learn useful representations from unlabeled data,  unsupervised learning methods, sometimes also known as self-supervised learning methods, have drawn significant attention. 

The general pipeline of unsupervised learning with a deep neural network is to design a "pretext" task for the network to solve while the label for this pretext tasks can be automatically generated based on the attributes of the data. After the network is trained with the pretext task, the network will be able to capture useful features. Recently, many unsupervised learning methods have been proposed to learn image features by training networks to solve pretext tasks, such as playing image jigsaw \cite{JigSaw}, clustering images \cite{deepcluster}, predicting image rotations \cite{RotNet}, image inpainting \cite{ContextEncoder}, generating images with generation adversarial network \cite{DCGAN}, etc. The unsupervised learning methods for image feature learning have obtained great success and the performance of unsupervised learning methods sometimes come very close to supervised methods \cite{deepcluster,selfsurvey}.

A number of unsupervised learning methods have also been proposed for point cloud unsupervised learning \cite{SPH, LFD, TLNet, 3DGAN, LatentGAN, FoldingNet}. Most of them are based on auto-encoders \cite{TLNet, 3DGAN, LatentGAN, FoldingNet}. Various auto-encoders are proposed to obtain features by training them to reconstruct the 3D point cloud data. Since the main purpose of such auto-encoders is to reconstruct the data, the networks thus trained may memorize the low-level features of the point cloud.

In this paper, we propose an unsupervised feature learning approach for point cloud by training GCNNs to solve two pretext tasks consecutively, which are part contrasting and object clustering.
Specifically, the network is trained to accomplish two pretext tasks: to compare (contrast) two point cloud cuts and to cluster point cloud objects. First, all the 3D point objects are cut into two parts and a GCNN (called ContrastNet) is trained to verify whether two randomly sampled parts from the dataset belong to the same object. Second, the point cloud data is clustered into clusters by using the features learned by the ContrastNet, and another GCNN (called ClusterNet) is trained to predict the cluster ID of each point cloud data. 

In summary, our main contributions in this paper are as follows:

\begin{itemize}
 
  \item{A simple and effective unsupervised feature learning framework is proposed for point cloud data. By training deep graph CNNs to solve two pretext tasks, part contrasting and object clustering, the networks are able to learn semantic features for point cloud data without using any annotations.} 
  
  

  \item{Extensive experiments are performed showing that our proposed approach outperforms most of the state-of-the-art unsupervised learning methods. With the proposed unsupervised method, our model obtains $86.8$\% and $93.8$\% classification accuracy on ModelNet40 and ModelNet10 datasets respectively.}

  \item{As a practical consideration, we have also tested object recognition using partial 3D data, by simulating occlusions and perspective views. Experiments show that our proposed approach generates results that are practically useful.}

 \end{itemize}

\section{Related Work}

\textbf{Point Cloud Understanding:} Various approaches have been proposed for point cloud understanding tasks, including classification, segmentation, and recognition and detection.  These approaches can be classified into three types: hand-crafted methods \cite{hancCraft1, hancCraft2, hancCraft3, hancCraft4, hancCraft5, hancCraft6, hancCraft7, LFD}, CNNs on regular 3D data \cite{volum1, modelnet, volum3, volum4, volum5, multiview1, multiview2, feat1, feat2}, and CNNs on unordered 3D point cloud data \cite{PointNet, PointNetPlusPlus, DGCNN, pointcnn}.

The first type of methods is hand-crafted based methods. These traditional methods capture the local geometric structure information of point cloud data such as intrinsic descriptors \cite{hancCraft1, hancCraft2, hancCraft3}, or extrinsic descriptors \cite{hancCraft4, hancCraft5, hancCraft6, hancCraft7, LFD}. These methods have very limited performance of 3D data analysis. 

\begin{figure*}[!ht]
\begin{center}
\includegraphics[width=\textwidth]{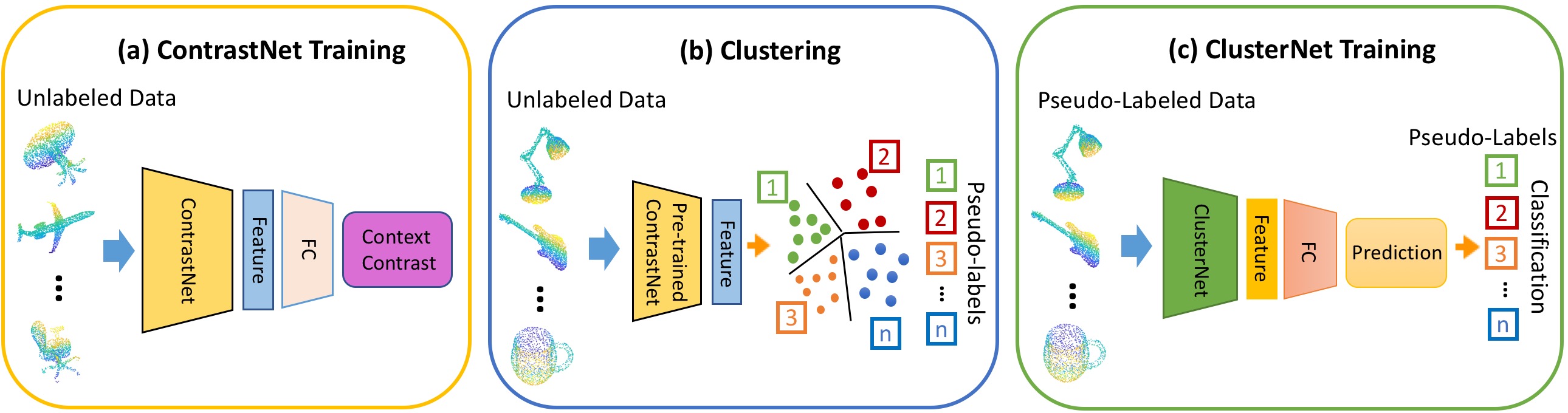}
\end{center}
\caption{The unsupervised feature learning pipeline includes three main steps: (a) ContrastNet for part contrast learning, by verifying whether two point cloud cuts belong to the same object; (b) Cluster samples of 3D objects and assign cluster IDs, using the features learned by ContrastNet; (c) ClusterNet for object clustering learning, by training the network with the 3D point cloud data while the labels are the cluster IDs assigned by the clustering step.}
\label{fig:Pipline}
\end{figure*}

Applying 3D convolutional neural networks to regular 3D data usually obtained better performance than traditional hand-crafted features. There are several approaches to handle the regular 3D data with CNNs: volumetric methods \cite{volum1, modelnet, volum3, volum4, volum5} voxelize unordered data to a static 3D grid then 3D CNNs are used to process the data. This kind of methods has a constraint on efficiency and complexity due to the data sparsity and cost of CNNs. Multi-view methods \cite{multiview1, multiview2} use 2D CNNs after rendering the 3D data into 2D images, which have obtained significant performance improvement on the classification task. However, this kind of methods has constraints in doing point level task, such as segmentation. Spectral methods \cite{spectral1, spectral2} apply spectral CNNs on meshes that are constrained by the expandability to other data formats. Feature-extracting methods \cite{feat1, feat2} extract features of 3D data and then apply CNNs to the features, which deeply depend on the quality of the extracted features.

Recently, a number of methods have been proposed for understanding unordered point cloud data \cite{PointNet, PointNetPlusPlus, DGCNN, pointcnn}. Qi \textit{et al.} made the first attempt to design a deep network architecture, named PointNet \cite{PointNet}, for using unordered point cloud to perform 3D shape classification, shape part segmentation and scene semantic parsing tasks. PointNet process each 3D point in a sample individually, therefore disarrangement of the point cloud will not constrain the function of the model. However, because of this, PointNet does not utilize the local structure of point cloud, which limits its ability to recognize fine-grained patterns. Later, they proposed  PointNet++ which applied PointNet recursively on a nested partitioning of the input point set \cite{PointNetPlusPlus} to improve the PointNet and address the impact of local information lost. PointCNN \cite {pointcnn} was proposed as a generalization of typical CNNs to feature learning from point clouds by the permutation of the points into a latent and potentially canonical order and the weighting of the input features associated with the points. To capturing local structure, Wang \textit{et al.} proposed DGCNN \cite{DGCNN} with an edge-convolution network (EdgeConv) to specifically model local neighborhood information by applying convolutions over the k nearest neighbors calculated by KNN in metric space, and the k nearest neighbors can be dynamically updated in different layers.

\textbf{Unsupervised Feature Learning:} Various unsupervised learning methods have been proposed to learn features from unlabeled data \cite{SPH, LFD, TLNet, 3DGAN, LatentGAN, FoldingNet, deepcluster}. Girdhar \textit{et al.} proposed the TL-embedding network\cite{TLNet}, which consists of an autoencoder that ensures the representation is generative and a convolutional network that ensures the representation is predictable. Sharma \textit{et al.} proposed a fully convolutional volumetric autoencoder to learn volumetric representation from noisy data by estimating the voxel occupancy grids \cite{3DGAN}. Achlioptas \textit{et al.} proposed LatentGAN by introducing a new deep auto-encoder network with state-of-the-art reconstruction quality and generalization ability for point cloud data \cite{LatentGAN}. Yang \textit{et al.} proposed FoldingNet which is an end-to-end autoencoder that is the state-of-the-art for unsupervised feature learning on point clouds \cite{FoldingNet}. In their work, a graph-based enhancement is applied to the encoder to enforce local structures on top of PointNet, and a folding-based decoder deforms a canonical 2D grid onto the underlying 3D object surface of a point cloud. 

Most of the deep learning based methods use auto-encoder variations for learning features on unlabeled point cloud data. However, the purpose of the autoencoder is to reconstruct the data and the feature may have a good performance on low-level tasks such as completion, reconstruction, and denoise, but have an inferior performance on tasks demands more high-level semantic meanings. Therefore, we propose the ContrastNet and ClusterNet to learn features by exploring high-level semantic features.
Our method outperforms most of the unsupervised methods on two MoldelNet datasets and only $0.6$\% lower than the supervised method PointNet on the ModelNet40 dataset.

\section{Method}

\begin{figure*}[!ht]
\begin{center}
\includegraphics[width=\textwidth]{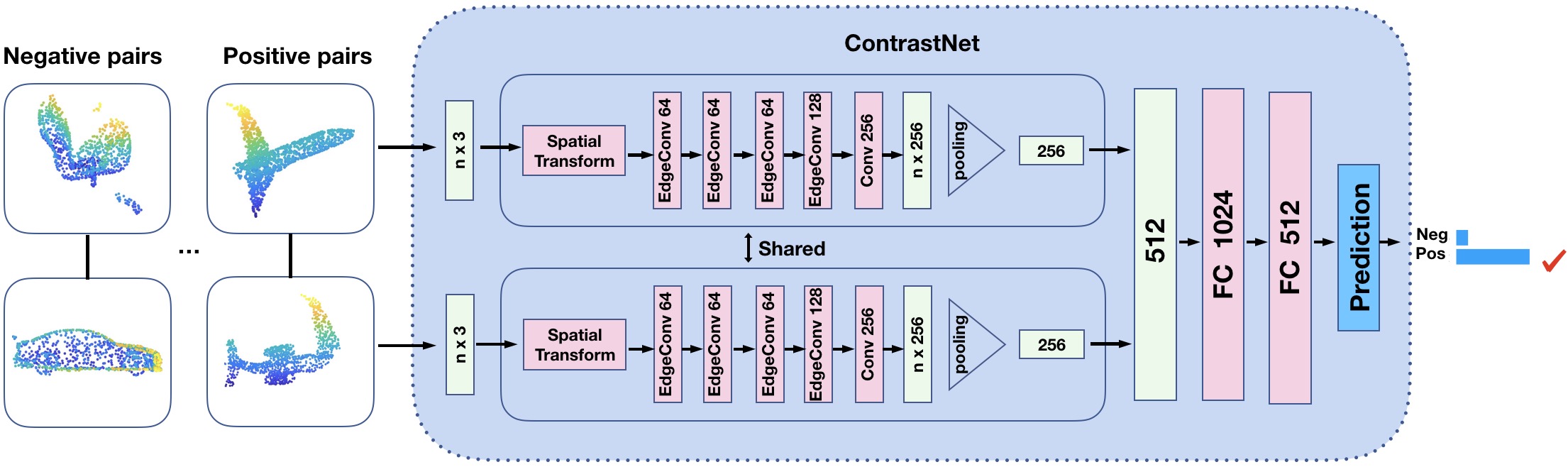}
\end{center}
\caption{The architecture of ContrastNet for part contrast learning. The positive pairs are generated by randomly sampling two segments from the same point cloud sample, while the negative pairs are generated by randomly sampling two segments from two different samples. A dynamic graph convolutional neural network (DGCNN)\cite{DGCNN} is used as the backbone network. Note that the top and bottom parts of the ContrastNet is the same DGCNN (i.e., sharing the same parameters). The features of two segments are concatenated and fed to fully connected layers to make the prediction of positive or negative. The part contrast learning does not require any data annotations by humans.}
\label{fig:ContextContrast}
\end{figure*}


To learn features from unlabeled point cloud data, we propose to learn features by training networks to accomplish both of the part contrasting and the object clustering pretext tasks. The pipeline of our framework is illustrated in Fig.~\ref{fig:Pipline}, which includes three major steps: ContrastNet for part contrast learning, clustering using the learned features, and then ClusterNet for object cluster learning using the cluster IDs. Here is a summary of the three modules before we get into details of the ContrastNet and ClusterNet.

\textbf{a) ContrastNet: Part Contrast Learning:} The first step is to learn features by training a network called ContrastNet to accomplish the part contrast task. Specifically, the part contrast task is to verify whether two point cloud segments (parts) belong to the same sample (object). The positive pair is drawn by selecting two different segments from the same object, while the negative pair is drawn by selecting two segments from two different objects. 

\textbf{b) Clustering to Obtain Pseudo-labels:}
After the training with the part contrasting finished, the trained ContrastNet can obtain high-level semantic features from point cloud data. Using the extracted features, the 3D point cloud data samples are clustered into different clusters. Kmeans++ \cite{kmeans} is used as the clustering algorithm in the paper. The point cloud data from the same cluster have high similarity while the data from different clusters have low similarity.

\textbf{c) ClusterNet: Classification using Pseudo-labels:}
Once obtained the clusters for the training data by using the Kmeans++ algorithm, the cluster IDs can be used as the pseudo-labels to train another network called ClusterNet. The clustering is used to boost the quality of the learned features of the ContrastNet. The architecture of the network for this step does not depend on the previous self-supervised model ContrastNet and therefore it can be flexibly designed as the demands.

\subsection{ContrastNet: Part Contrast Learning}

When a point cloud data is observed from different views, only part of the 3D object can be seen. The observable part can be very different based on the view. For example, as shown in Fig.~\ref{fig:part}, for the same airplane, when it is observed from different views, the observed segments can be totally different. However, the different segments still belong to the same object. 

Inspired by this observation, we proposed the part (segment) contrast as the first pretext task for a GCNN to solve. The task is defined as to train a ContrastNet to verify whether two point cloud segments belong to the same object. The positive pair is drawn by selecting two different segments from the same object, while the negative pair is drawn by selecting two segments from two different objects. The illustration of the part contrast task is shown in Fig.~\ref{fig:ContextContrast}.

We randomly split one object into two segments, thus generating a "part" dataset. Then a pair for the segments are randomly selected. If a pair of the segments are from the same object, this pair is a positive instance that will be labeled as 1. Otherwise, if a pair comes from two objects, it is a negative instance and will be labeled as 0. More implementation details will be introduced in Section~\ref{subsection:implementation}.
We model this task as a binary classification problem. As the training goes on, the segments from the same object should have a smaller distance while the segments from different objects have a larger distance. In this way, the semantic features can be learned.

Note that since a pair of parts from two objects that belong to the same category will be treated as a "negative" instance instead of "positive", the training of positive and negative has certain percentage of "error" in the input data. For example, in ModelNet40 dataset, objects belong to 40 categories. Without using the labels in training ContrastNet, approximately there is an 1/40 (2.5\%) error in the input data for verifying positive or negative instances.

As for the network architecture,  we choose DGCNN \cite{DGCNN} as the backbone model since this model specifically captures the local structure of the point cloud with dynamically constructed graphs and yields better performance. The details of the network architecture are shown in Fig.~\ref{fig:ContextContrast}. There are two branches, one for each point cloud segment, from a pair of input segments. Each branch consists of a spatial transformer network to align the point cloud and followed by $4$ EdgeConv layers (to construct graphs over k nearest neighbors calculated by KNN) with $64, 64, 64, 128$ kernel sizes, respectively. After which, one convolutional layer with $256$ channels is used to embed the four embeddings obtained by the four EdgeConv layers to high dimensional space. The feature then is pooled into a $256$-dimension vector by applying the max-pooling layer. The two feature vectors from the two branches are then concatenated into a vector to be fed to three fully-connected layers (with $1,024, 512, 2$  vector lengths, respectively). The ReLU activation and batch normalization are used for each layer and $50$\% dropout is used on each fully-connected layer. The cross-entropy loss is optimized by Adam and backpropagation to compute the gradient.

\subsection{ClusterNet: Knowledge Transfer with Clusters}
The underline intuition of clustering is that 3D objects from the same categories have high similarity than those from different categories. After obtaining the clusters of the data by using the Kmeans++, based on the features extracted by ContrastNet, the cluster IDs of the data are used as the "pseudo" labels to train a ClusterNet, so that more meaningful features may be extracted from it. We hope that using cluster IDs as pseudo labels in ClusterNet can provide more powerful self-supervision and therefore, the network can learn more representative features for object classification.

Given any unlabeled point cloud dataset $X = \{x_1, x_2, \dots , x_N\}$ of $N$ images, the clustering process can be parameterized as \cite{kmeans}: 
\begin{equation}
\label{eq:kmeans}
  \min_{C \in \mathbb{R}^{d\times k}}
  \frac{1}{N}
  \sum_{n=1}^N
  \min_{y_n \in \{0,1\}^{k}}
  \| f_\theta(x_n) -  C y_n \|_2^2,
  \quad
  \quad
  y_n^\top 1_k = 1,
\end{equation} where $f_\theta$ is the feature extractor that can map any point cloud data into a vector, $\theta$ is the set of corresponding parameters that need to be optimized, $y_n$ is the cluster ID. Solving this clustering problem provides a set of optimal assignments $(y_n^*)_{n\le N}$ and a centroid matrix $C^*$. The cluster ID assignments $(y_n^*)_{n\le N}$ are then used as the pseudo-labels to train a ClusterNet.

The training of the ClusterNet, also based on DGCNN \cite{DGCNN},  with the cluster ID assignments as the pseudo-labels, is described as:
\begin{eqnarray}\label{eq:sup}
  \min_{\theta, W} \frac{1}{N} \sum_{n=1}^N \ell\left(g_W\left( f_\theta(x_n) \right), y_n\right),
\end{eqnarray}
where the purpose of training is to find the optimal parameters $\theta^*$ such that the mapping $f_{\theta^*}$ produces good general-purpose features for point cloud data classification. A parameterized classifier $g_W$ predicts the correct labels of the data based on the features $f_\theta(x_n)$. All the parameters are learned by optimizing this loss function. In supervised training, parameters $\theta$ are optimized with the human-annotated labels while each data $x_n$ is paired with a human-annotated label $y_n$ in $\{0,1\}^k$. In our unsupervised learning training, each data $x_n$ is paired with a pseudo label $y_n$ that is generated by the clustering algorithm. The label $y_n$ indicates the data's membership to one of the $k$ clusters, where $k$ can be specified in the clustering algorithm. In our experiments (below), various number of clusters are tested for comparing the impact on feature extraction. 

\section{Experimental Results}
We conduct extensive experiments to evaluate the proposed approach and the quality of the learned features for point cloud on the point cloud classification task.

\subsection{Implementation Details} \label{subsection:implementation}

\textbf{ContrastNet:} During the part contrast unsupervised learning, each object is cut by randomly generated $15$ planes into $30$ segments. Each selected segment has at least $512$ points. Any two segments from the same object are treated as the positive samples while any two segments from two different objects are treated as negative samples. The DGCNN is used as the backbone of the ContrastNet. During the unsupervised part contrast training phase, the learning rate is $0.001$, momentum is $0.9$, the optimizer is Adam, the learning rate decay rate is $0.7$, and the decay step is $200000$. 

\textbf{ClusterNet:} The Kmeans++ is used as the clustering algorithm to cluster the data based on the embeddings extracted by the ContrastNet. We test the performance of different cluster numbers to train the ClusterNet. The same DGCNN  structure is used as the backbone in the ClusterNet except that the size of the last dense layer is the cluster number. During the training with pseudo labels, the learning rate is $0.001$, momentum is $0.9$, the optimizer is Adam, the learning rate decay rate is $0.7$, and the decay step is $200000$.

\subsection{Datasets}
All the experiments for both the ContrastNet and ClusterNet are done on three point cloud benchmarks: ModelNet40, ShapeNet, and ModelNet10. Data augmentation including random rotation, shift, and jittering are used during all the training phases. 
The \textbf{ModelNet40} dataset contains $12,311$ meshed CAD models covering $40$ classes. There are $9,843$ and $2,468$ samples in the training and testing splits, respectively. In all our experiments, $1024$ points are randomly picked for each model during the training and testing phases. This dataset is used to train and test our unsupervised learning method. During training, this dataset has been used for learning features without using the class labels. During the testing phase, this dataset is used to evaluate the quality of the learned features. 
The \textbf{ModelNet10} dataset contains 10 categories including 3991 meshed CAD models for training and 909 models for testing. 
We randomly sample $2048$ points from the mesh faces and use their (x, y, z) coordinates as input in all experiments. This model is only used for testing the quality of the learned features.
The \textbf{ShapeNet} part dataset that contains 16 categories including $12,137$ models for training and $2,874$ for testing from ShapeNet dataset. In all our experiments, $1024$ points are randomly picked for each model during the training and testing phases. This dataset is used for unsupervised training.

\subsection{Can ContrastNet Fulfill Part Contrast Task?}

The hypothesis of our idea is that the ContrastNet is able to learn semantic features by accomplishing the "part contrasting" pretext task, and then the learned features can be used for other downstream tasks, such as point cloud classification. Therefore, we test the performance of ContrastNet in verifying whether two patches belong to the same object. No human-annotated labels are used during the training phase of the ContrastNet. 

The performance of the ContrastNet in part contrasting is shown in Table~\ref{tab:contextcontrast}. The average accuracy of part contrasting is more than $90$\% on the two datasets, with cross-dataset testing.

\begin{table}[hb]
\begin{center}
\begin{tabular}{l|c|c}
\hline
Training & Testing & Accuracy(\%) \\
\hline\hline
ShapeNet     & ShapeNet  & 94.0\\
ShapeNet     & ModelNet40  & 86.4 \\
ModelNet40    & ModelNet40 & 95.0 \\
ModelNet40    & ShapeNet & 90.9 \\
\hline
\end{tabular}
\end{center}

\caption{Performance of ContrastNet in part contrasting on ShapeNet and ModelNet40 datasets.} 
\label{tab:contextcontrast}
\end{table}

\begin{figure}[!ht]
\begin{center}
\includegraphics[width=0.5\textwidth]{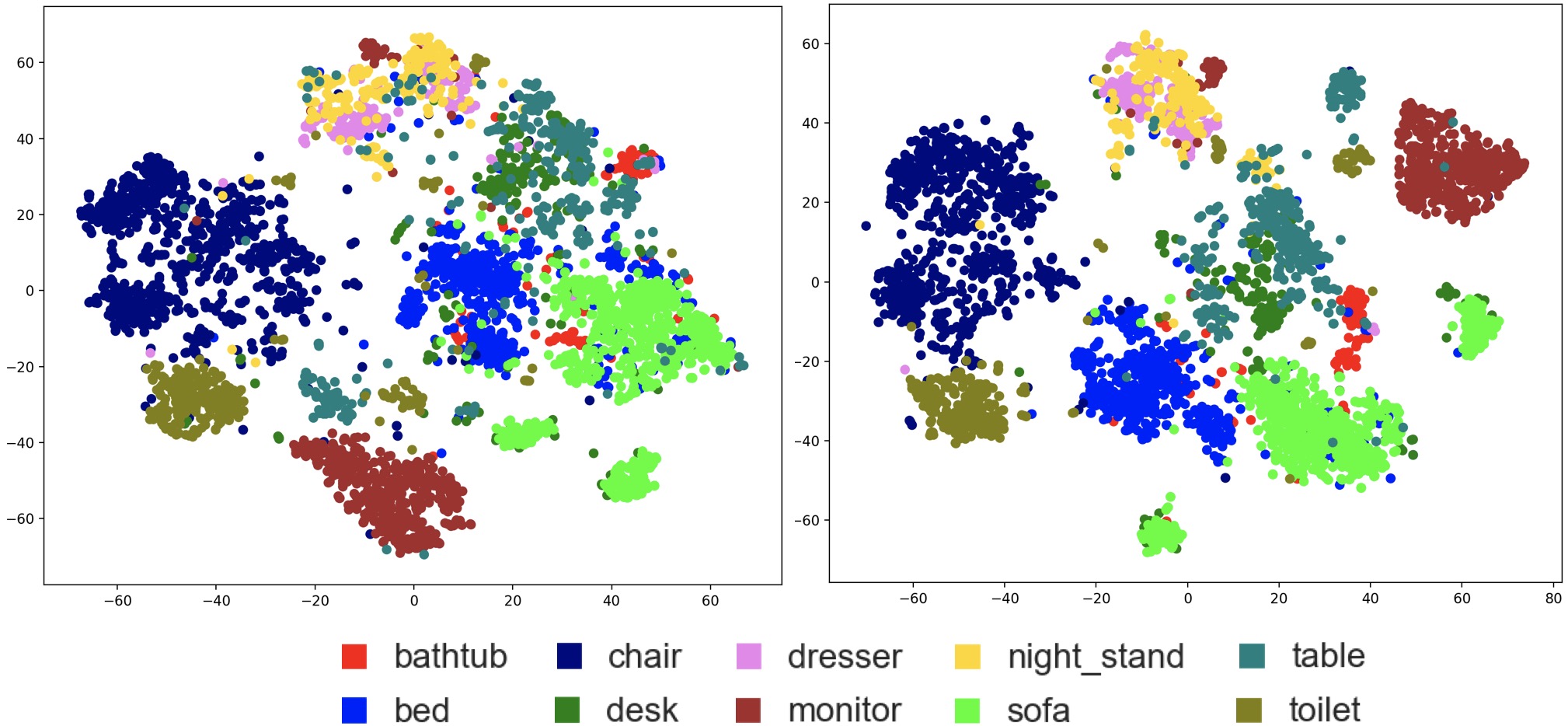}
\end{center}
\caption{Visualization of object embedding of the ModelNet10 test data through part contrast training on the ShapeNet dataset. The features are learned by part contrast learning (left) and then boosted by object clustering (right).}
\label{fig:tsne}
\end{figure}

To verify whether the ContrastNet learned useful features, we visualize the testing data by using TSNE, as shown in Fig.~\ref{fig:tsne} (left).  All of the data covering 10 classes of ModelNet10 is visualized. The figure shows that the ContrastNet indeed learned semantic features and the data from the same class are closer than the data from different classes. Later Clustering enhances the features more (as shown in Fig.~\ref{fig:tsne} (right), which will be discussed more in the impact on classification.

\subsection{Transfer Features Learned by ContrastNet to Classification Task}

To quantitatively evaluate the quality of the learned features for classification by using the part contrasting pretext task, we conduct experiments on three different datasets:  ShapeNet, ModelNet10, and ModelNet40. The features are extracted by the ContrastNet only trained for the part contrasting task on unlabeled data. A linear classifier SVM is trained based on the learned features, and the testing classification accuracy is reported in the column "ContrastNet" in Table~\ref{tab:cluster}.  
Following the practice of previous work \cite{FoldingNet, 3DGAN}, we conduct cross-dataset training and testing to verify the generalization ability of features between different datasets. As shown in Table~\ref{tab:cluster}, when trained only with a linear classifier SVM on one dataset, the ContrastNet trained on ShapeNet is able to achieve $84.1$\% and $91.0$\% on ModelNet40 and ModelNet10 dataset respectively. As a comparison, the model trained on ModelNet40 and tested on the same dataset achieved $85.7$\%. These results validate the effectiveness of the proposed method and that the learned features by the proposed unsupervised learning method can be transferred among different datasets. 


\subsection{Can Clustering Boost Performance?}

The part contrast learning indeed forces the ContrastNet to learn semantic features. However, the fact that objects belong to the same classes were treated as different objects in part contrasting (without knowing their labels) may have a negative impact on the quality of learned features by ContrastNet. Therefore, clustering is applied to discover the objects with similar appearances and the ClusterNet is trained to learn features by using the cluster IDs as object labels. We hope that the ClusterNet should be able to help the model learn more discriminative features for classification. 

The features extracted by the ContrastNet for a dataset are used to group the data into a number of clusters using the Kmeans++ algorithm.  
A ClusterNet is trained from scratch with 3D point cloud data as the input to predict the cluster ID of each data sample.
After the training is finished, the network is tested on the point cloud classification task using the same SVM on the same three benchmarks as above. The classification results are shown in column "ClusterNet" of Table~\ref{tab:cluster}, where the number of clusters is selected as 300 (see below for a discussion).

\begin{table}[!hb]
\begin{center}
\small
\begin{tabular}{l|c|c|c}
\hline
Training & Testing & ContrastNet(\%) & ClusterNet(\%) \\
\hline\hline
ShapeNet& ModelNet40  & 84.1 &86.8 (\textbf{+2.7})\\
ShapeNet   & ModelNet10  & 91.0 &93.8 (\textbf{+2.8})\\
ModelNet40  & ModelNet40 & 85.7 &88.6 (\textbf{+2.9})\\
\hline
\end{tabular}
\end{center}
\caption{Comparison of 3D object classification results using ContrastNet and ClusterNet. } 
\label{tab:cluster}
\end{table}

As shown in Table~\ref{tab:cluster}, training the ClusterNet to predict the cluster ID of each data, generated by Kmeans++ based on the features learned by ContrastNet, can significantly boost the point cloud classification accuracy. The clustering boosts the classification accuracy on all the three datasets by at least $2.7$\%. These improvements validate the effectiveness of using clustering to boost the quality of the learned features, also visualized shown in Fig.~\ref{fig:tsne}.

We also conduct experiments to evaluate the impact of the numbers of clusters on the quality of features. By varying the numbers of clusters, we have examined the point cloud classification accuracy on the three benchmark datasets.
As shown in Table~\ref{tab:clusterRes}, the points cloud classification performance first improved when larger cluster numbers are used and then saturated when the numbers are larger than certain values. When more clusters are applied, the fine-grained object groups should be discovered which probably leads to more discriminative features.

\begin{table}[hb]
\begin{center}
\begin{tabular}{c|c|c|c}
\hline
            & ShapeNet & ShapeNet & ModelNet40 \\
Clusters  & ModelNet40 & ModelNet10 & ModelNet40 \\
\hline\hline
100    & 86.2\%   & 93.5\%  & 87.7\% \\
200    & 86.4\%   & 93.6\%   & 88.2\%\\
300    & \textbf{86.8\%}  & \textbf{93.8\%}   & \textbf{88.6\%} \\
400    & 86.5\%   & 93.2\%   & 87.8\% \\
\hline
\end{tabular}
\end{center}
\caption{The relation of number of clusters and the performance on point cloud classification. The classification performance improved slightly when larger cluster numbers are used.}
\label{tab:clusterRes}
\end{table}

\subsection{Quality of Clustering: Further Study}
To further analysis the quality of learned features, we evaluate the quality of the clusters by calculating the accuracy of each cluster. Specifically, for each cluster, we assign the category label of the majority data as the label and evaluate the accuracy of all the data. We cluster the ShapeNet and ModelNet40 into $16$ and $40$ clusters respectively, since these numbers equal to the actual numbers of categories in the two datasets. 

As shown in Table~\ref{tab:clusterAcc}, the cluster accuracy on ShapeNet is $83.4$\% which means that $83.4$\% of the data are correctly clustered into the same labels using Kmeans++.  The clustering accuracy on ModelNet40 is $64.2$\%, which is much lower than that of ShapNet, probably because ModelNet has more categories than ShapeNet.  Nevertheless, even with the low clustering accuracy, the testing accuracy obtained after the ClusterNet using SVM on ModelNet40 dataset is $87.4$\%, a $23.2$\% "improvement" over the training data "accuracy", which means ClusterNet can significantly optimize the quality of the features with a clustering step for the data.

\begin{table}[!hb]
\begin{center}
\begin{tabular}{l|c|c|c}
\hline
Pre-training  & Clusters & Clustering Acc. & Testing Acc. \\
\hline\hline
ShapeNet      & 16   & 83.4\%    & 86.1\% \\
ModelNet40    & 40   & 64.2\%    & 87.4\% \\
\hline
\end{tabular}
\end{center}
\caption{The accuracy of the clustering and testing results. 
The clustering accuracy seems to depend on the numbers of the categories. The testing accuracy are obtained from ClusterNet using SVM, that has $23.2$\% improvement on ModelNet40.}
\label{tab:clusterAcc}
\end{table}

\begin{figure}[!ht]
\begin{center}
\includegraphics[width=0.40\textwidth]{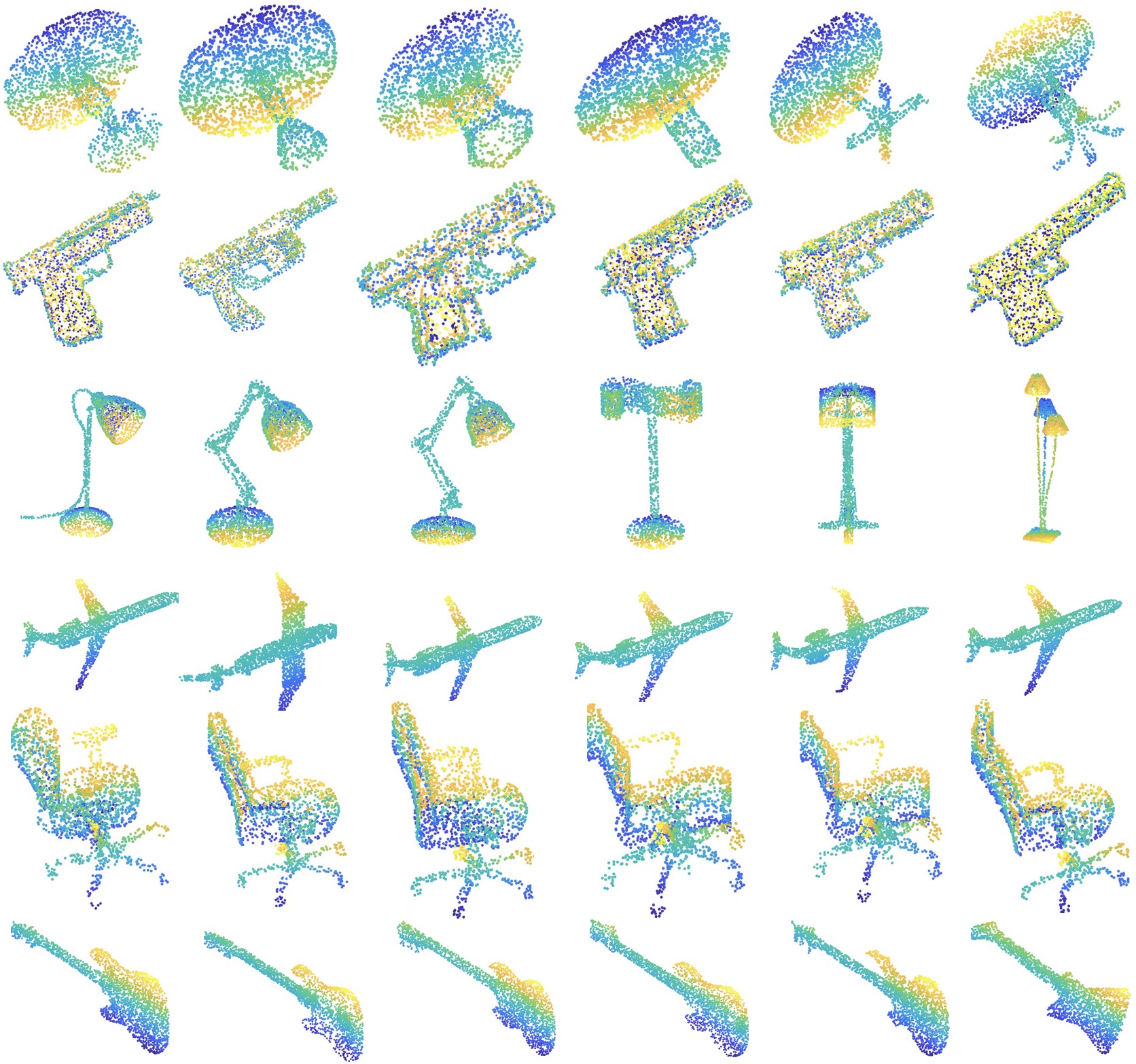}
\end{center}
\caption{Visualizing clustering result after applying Kmeans++ on the unlabeled data. The first column is the discovered centrioid of each cluster and the other five columns are the top five closest data to the centroid. }
\vspace{-10pt}
\label{fig:cluster}
\end{figure}

The clustering can cluster point cloud objects into groups that the objects from the same groups have smaller distances in the feature space while objects from different groups have larger distances in the feature space. The quality of the cluster indicates the discriminative ability of the learned features. Therefore, we randomly select $6$ clusters and show the cluster center of each and the top $5$ objects that closest to the cluster center. 
As shown in Fig.~\ref{fig:cluster}, the object from the same cluster have very high similar appearance and geometry. 

\subsection{Practical Considerations: Occlusions and Perspective Views}

\begin{figure}[!ht]
\begin{center}
\includegraphics[width=0.45\textwidth]{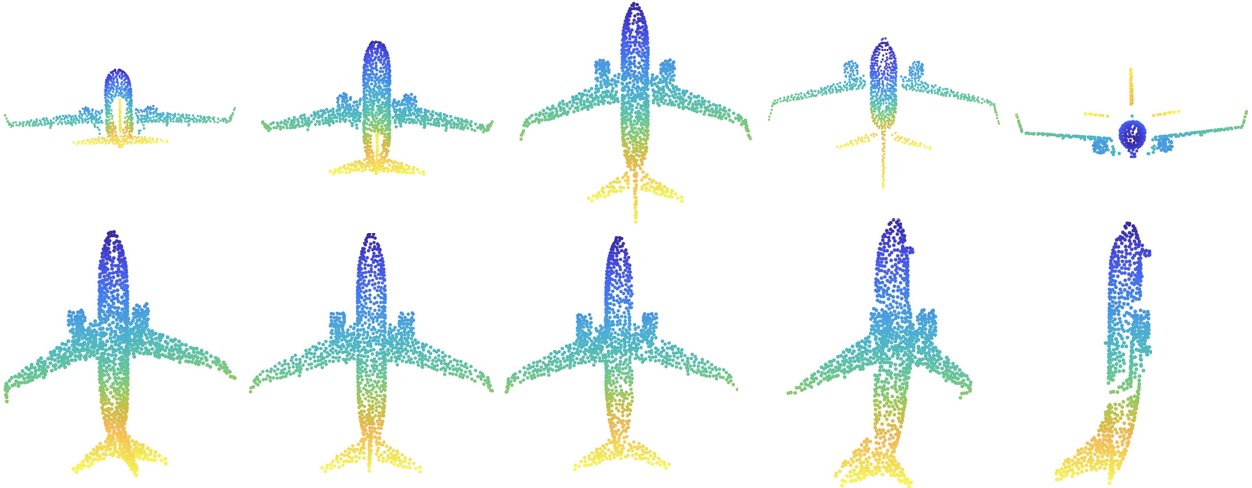}
\end{center}
\caption{Several perspective views of 3D point cloud of an object. The views from different perspectives might be totally different. }
\vspace{-5pt}
\label{fig:pers}
\end{figure}

 We have known that ContrastNet and ClusterNet can learn useful features for recognizing  full objects. However, in real-life a sensor can only observe part of an object due to occlusions and/or perspective views.  Fig.~\ref{fig:pers} and Fig.~\ref{fig:part} show that different perspective views and various part segments (respectively) from one object might be very different from each other, even when 3D point cloud can still be obtained. To verify if our deep models can still classify part segments and perspective views, we train two sets of our ContrasNet and ClusterNet models and test the classification accuracy on ModelNet40, using part segments and perspective views, respectively.  In each case, after extracting features using the ContrasNet using the part segments (or perspective views), the pseudo labels are obtained using the ContrastNet features with Kmeans++. Then we train a ClusterNet for each case and extract features using part segments (or perspective views) on ModelNet40. Note that in each case, the features in both training and testing and by both the ContrastNet model and ClusterNet model are extracted on part segments (or perspective views) instead of full objects. Each part segment (or perspective view) shall contain at least $512$ points. A linear SVM is trained based on the features extracted from the two steps (ContrastNet and ClusterNet) to obtain the classification performance. 

The results in Table~\ref{tab:partAcc} shows that the accuracy (in column "Acc. Part") of part segments classification and the accuracy (in column "Acc. Perspective") of perspective views classification are $82.4$\% and $75.8$\%. This indicates that even though we only feed the part segments or perspective views to our networks, they still can learning  high-quality features for the classification task. The results validate the practical usefulness of the proposed approach. Nevertheless, we would like to note that when the full object data points are used for the training and testing (in column "Acc. Full", listed again here from Table~\ref{tab:cluster} ), the performance is around $6$\% higher than using the features obtained from part segments and around $13$\% higher than using the features from perspective views. This might also indicate that the perspective views which only have 3D points on surfaces from single perspective views are not as descriptive as parts of the volumetric body of a 3D object.

\begin{table}[!hb]
\begin{center}
\begin{tabular}{l|c|c|c}
\hline
Model & Acc. Full & Acc. Part & Acc. Perspective\\
\hline\hline
ContrastNet & 85.7\% & 79.4\% & 72.0\% \\
ClusterNet & 88.6\% & 82.4\% & 75.8\% \\
\hline
\end{tabular}
\end{center}
\caption{The columns "Acc. Full", "Acc. Part", and "Acc. Perspective" are accuracy obtained from the features extracted on full objects, part segments, and perspective views, respectively. }
\vspace{-5pt}
\label{tab:partAcc}
\end{table}

\subsection{Comparison with the State of the Art}
In this section, we compare our approach with both supervised models \cite{PointNet, PointNetPlusPlus, pointcnn,DGCNN} and other unsupervised learning models \cite{SPH, LFD, TLNet, Vconvdae, 3DGAN, LatentGAN, FoldingNet} on point cloud classification benchmarks  ModelNet10 and ModelNet40. In all comparison, we use the ClusterNet.

Following the common practice \cite{FoldingNet, 3DGAN}, all the unsupervised models are trained on the ShapeNet data with the same procedure (Table~\ref{tab:compare}). All the methods ran a linear SVM upon on the high-dimensional features obtained by usnupervised training. The methods in \cite{SPH, LFD} are hand-crafted features and methods in \cite{TLNet, Vconvdae, 3DGAN, LatentGAN, FoldingNet} are deep learning based methods.  On the MoldelNet40 dataset, our method outperforms all the methods except FoldingNet  ($1.6$\% lower), which is the latest work for unsupervised feature learning. On the ModelNet10 dataset, our methods outperforms SPH \cite{SPH}, LFD \cite{LFD}, TLNetwork \cite{TLNet}, VConv-DAE \cite{Vconvdae}, and 3DGAN \cite{3DGAN}, and only $0.6$\% lower than FoldingNet \cite{FoldingNet}.  We would like to note that our ClusterNet has a much simpler structure and is much easier in training.

\begin{table}[!h]
\begin{center}
\begin{tabular}{l|c|c}
\hline
Models & ModelNet40 (\%) & ModelNet10 (\%) \\
\hline\hline
SPH \cite{SPH} & 68.2  & 79.8\\
LFD \cite{LFD} & 75.5  & 79.9 \\
T-L Network \cite{TLNet} & 74.4  & - \\
VConv-DAE \cite{Vconvdae} & 75.5  & 80.5 \\
3D-GAN \cite{3DGAN} & 83.3  & 91.0 \\
Latent-GAN \cite{LatentGAN} & 85.7 & \textbf{95.3} \\
FoldingNet \cite{FoldingNet} & \textbf{88.4}  & 94.4 \\
\textbf{\textit{ClusterNet (Ours)}} & \textbf{\textit{86.8}} & \textbf{\textit{93.8}} \\
\hline
\end{tabular}
\end{center}
\caption{The comparison on classification accuracy between our ClusterNet and other unsupervised methods on point cloud classificaton dataset ModelNet40 and ModelNet10. } 
\label{tab:compare}
\vspace{-5pt}
\end{table}

We also compare the performance with the most recent supervised methods including PointNet \cite{PointNet}, PointNet++ \cite{PointNetPlusPlus}, PointCNN \cite{pointcnn}, and DGCNN \cite{DGCNN}. All the parameters of these methods are trained with human-annotated labels, while our results are obtained by training linear SVM based on the features extracted by the ClustertNet. 
As shown in Table~\ref{tab:compareSuper}, the supervised methods have better performance because all the parameters are tuned by the hand-annotated labels. With the unsupervised learned features and a linear SVM, the performance of our model (using unsupervised DGCNN as the base model) is only $3.6$\% lower than the supervised DGCNN. These results demonstrate the effectiveness of our unsupervised learning method.

\begin{table}[!h]
\begin{center}
\begin{tabular}{l|c|l|c}
\hline
Models  & Acc.(\%) & Models & Acc.(\%)  \\
\hline\hline
PointNet \cite{PointNet} & 89.2 & DGCNN \cite{DGCNN} & \textbf{92.2} \\
PointNet++ \cite{PointNetPlusPlus} & 90.7 & & \\
PointCNN \cite{pointcnn} & \textbf{92.2} &\textbf{\textit{ClusterNet(Ours)}} & \textbf{\textit{88.6}}\\
\hline
\end{tabular}
\end{center}
\caption{The comparison on classification accuracy between our unsupervised ClusterNet and the supervised methods on point cloud classificaton on ModelNet40.}
\label{tab:compareSuper}
\vspace{-10pt}
\end{table}

\section{Conclusion}
We have proposed a straightforward and effective method for learning high-level features for point cloud data from unlabeled data.  The experiment results demonstrate that proposed pretext tasks (part contrasting and object clustering) are able to provide essential semantic information of the point cloud data for the network to learn semantic features. Our proposed methods have been evaluated on three public point cloud benchmarks and obtained comparable performance with other state-of-the-art self-supervised learning methods and showed practical applications to occluded and perspective data. 

\section{Acknowledgments}
The work is supported by the US National Science Foundation via awards \#CNS-1737533 and \#IIP-1827505, and Bentley Systems, Inc. under a CUNY-Bentley Collaborative Research Agreement (CRA).

{\small
\bibliographystyle{ieee}
\bibliography{3DV)
}

\begin{thebibliography}{10}\itemsep=-1pt

\bibitem{LatentGAN}
P.~Achlioptas, O.~Diamanti, I.~Mitliagkas, and L.~Guibas.
\newblock Representation learning and adversarial generation of 3d point
  clouds.
\newblock {\em arXiv preprint arXiv:1707.02392}, 2(3):4, 2017.

\bibitem{hancCraft1}
M.~Aubry, U.~Schlickewei, and D.~Cremers.
\newblock The wave kernel signature: A quantum mechanical approach to shape
  analysis.
\newblock In {\em ICCVW}, pages 1626--1633. IEEE, 2011.

\bibitem{hancCraft3}
M.~M. Bronstein and I.~Kokkinos.
\newblock Scale-invariant heat kernel signatures for non-rigid shape
  recognition.
\newblock In {\em 2010 IEEE Computer Society Conference on Computer Vision and
  Pattern Recognition}, pages 1704--1711. IEEE, 2010.

\bibitem{spectral1}
J.~Bruna, W.~Zaremba, A.~Szlam, and Y.~LeCun.
\newblock Spectral networks and locally connected networks on graphs.
\newblock {\em arXiv preprint arXiv:1312.6203}, 2013.

\bibitem{deepcluster}
M.~Caron, P.~Bojanowski, A.~Joulin, and M.~Douze.
\newblock Deep clustering for unsupervised learning of visual features.
\newblock In {\em ECCV}, pages 132--149, 2018.

\bibitem{ShapeNet}
A.~X. Chang, T.~Funkhouser, L.~Guibas, P.~Hanrahan, Q.~Huang, Z.~Li,
  S.~Savarese, M.~Savva, S.~Song, H.~Su, et~al.
\newblock Shapenet: An information-rich 3d model repository.
\newblock {\em arXiv preprint arXiv:1512.03012}, 2015.

\bibitem{LFD}
D.-Y. Chen, X.-P. Tian, Y.-T. Shen, and M.~Ouhyoung.
\newblock On visual similarity based 3d model retrieval.
\newblock In {\em Computer graphics forum}, volume~22, pages 223--232. Wiley
  Online Library, 2003.

\bibitem{feat1}
Y.~Fang, J.~Xie, G.~Dai, M.~Wang, F.~Zhu, T.~Xu, and E.~Wong.
\newblock 3d deep shape descriptor.
\newblock In {\em CVPR}, pages 2319--2328, 2015.

\bibitem{TLNet}
R.~Girdhar, D.~F. Fouhey, M.~Rodriguez, and A.~Gupta.
\newblock Learning a predictable and generative vector representation for
  objects.
\newblock In {\em ECCV}, pages 484--499. Springer, 2016.

\bibitem{feat2}
K.~Guo, D.~Zou, and X.~Chen.
\newblock 3d mesh labeling via deep convolutional neural networks.
\newblock {\em TOG}, 35(1):3, 2015.

\bibitem{kmeans}
J.~A. Hartigan and M.~A. Wong.
\newblock Algorithm as 136: A k-means clustering algorithm.
\newblock {\em Journal of the Royal Statistical Society. Series C (Applied
  Statistics)}, 28(1):100--108, 1979.

\bibitem{RotNet}
L.~Jing and Y.~Tian.
\newblock Self-supervised spatiotemporal feature learning by video geometric
  transformations.
\newblock {\em arXiv preprint arXiv:1811.11387}, 2018.

\bibitem{selfsurvey}
L.~Jing and Y.~Tian.
\newblock Self-supervised visual feature learning with deep neural networks: A
  survey.
\newblock {\em arXiv preprint arXiv:1902.06162}, 2019.

\bibitem{hancCraft7}
A.~E. Johnson and M.~Hebert.
\newblock Using spin images for efficient object recognition in cluttered 3d
  scenes.
\newblock {\em TPAMI}, 21(5):433--449, 1999.

\bibitem{SPH}
M.~Kazhdan, T.~Funkhouser, and S.~Rusinkiewicz.
\newblock Rotation invariant spherical harmonic representation of 3 d shape
  descriptors.
\newblock In {\em Symposium on geometry processing}, volume~6, pages 156--164,
  2003.

\bibitem{volum4}
R.~Klokov and V.~Lempitsky.
\newblock Escape from cells: Deep kd-networks for the recognition of 3d point
  cloud models.
\newblock In {\em ICCV}, pages 863--872, 2017.

\bibitem{pointcnn}
Y.~Li, R.~Bu, M.~Sun, W.~Wu, X.~Di, and B.~Chen.
\newblock Pointcnn: Convolution on x-transformed points.
\newblock In {\em Advances in Neural Information Processing Systems}, pages
  820--830, 2018.

\bibitem{hancCraft6}
H.~Ling and D.~W. Jacobs.
\newblock Shape classification using the inner-distance.
\newblock {\em TPAMI}, 29(2):286--299, 2007.

\bibitem{spectral2}
J.~Masci, D.~Boscaini, M.~Bronstein, and P.~Vandergheynst.
\newblock Geodesic convolutional neural networks on riemannian manifolds.
\newblock In {\em ICCVW}, pages 37--45, 2015.

\bibitem{volum1}
D.~Maturana and S.~Scherer.
\newblock Voxnet: A 3d convolutional neural network for real-time object
  recognition.
\newblock In {\em IROS}, pages 922--928. IEEE, 2015.

\bibitem{JigSaw}
M.~Noroozi and P.~Favaro.
\newblock Unsupervised learning of visual representations by solving jigsaw
  puzzles.
\newblock In {\em ECCV}, pages 69--84. Springer, 2016.

\bibitem{ContextEncoder}
D.~Pathak, P.~Krahenbuhl, J.~Donahue, T.~Darrell, and A.~A. Efros.
\newblock Context encoders: Feature learning by inpainting.
\newblock In {\em CVPR}, pages 2536--2544, 2016.

\bibitem{PointNet}
C.~R. Qi, H.~Su, K.~Mo, and L.~J. Guibas.
\newblock Pointnet: Deep learning on point sets for 3d classification and
  segmentation.
\newblock In {\em CVPR}, pages 652--660, 2017.

\bibitem{volum5}
C.~R. Qi, H.~Su, M.~Nie{\ss}ner, A.~Dai, M.~Yan, and L.~J. Guibas.
\newblock Volumetric and multi-view cnns for object classification on 3d data.
\newblock In {\em CVPR}, pages 5648--5656, 2016.

\bibitem{PointNetPlusPlus}
C.~R. Qi, L.~Yi, H.~Su, and L.~J. Guibas.
\newblock Pointnet++: Deep hierarchical feature learning on point sets in a
  metric space.
\newblock In {\em NIPS}, pages 5099--5108, 2017.

\bibitem{DCGAN}
A.~Radford, L.~Metz, and S.~Chintala.
\newblock Unsupervised representation learning with deep convolutional
  generative adversarial networks.
\newblock {\em arXiv preprint arXiv:1511.06434}, 2015.

\bibitem{hancCraft5}
R.~B. Rusu, N.~Blodow, and M.~Beetz.
\newblock Fast point feature histograms (fpfh) for 3d registration.
\newblock In {\em ICRA}, pages 3212--3217. IEEE, 2009.

\bibitem{hancCraft4}
R.~B. Rusu, N.~Blodow, Z.~C. Marton, and M.~Beetz.
\newblock Aligning point cloud views using persistent feature histograms.
\newblock In {\em 2008 IEEE/RSJ International Conference on Intelligent Robots
  and Systems}, pages 3384--3391. IEEE, 2008.

\bibitem{multiview2}
M.~Savva, F.~Yu, H.~Su, M.~Aono, B.~Chen, D.~Cohen-Or, W.~Deng, H.~Su, S.~Bai,
  X.~Bai, et~al.
\newblock Shrec16 track: largescale 3d shape retrieval from shapenet core55.
\newblock In {\em Proceedings of the eurographics workshop on 3D object
  retrieval}, pages 89--98, 2016.

\bibitem{Vconvdae}
A.~Sharma, O.~Grau, and M.~Fritz.
\newblock Vconv-dae: Deep volumetric shape learning without object labels.
\newblock In {\em European Conference on Computer Vision}, pages 236--250.
  Springer, 2016.

\bibitem{SUNCG}
S.~Song, F.~Yu, A.~Zeng, A.~X. Chang, M.~Savva, and T.~Funkhouser.
\newblock Semantic scene completion from a single depth image.
\newblock In {\em CVPR}, pages 1746--1754, 2017.

\bibitem{multiview1}
H.~Su, S.~Maji, E.~Kalogerakis, and E.~Learned-Miller.
\newblock Multi-view convolutional neural networks for 3d shape recognition.
\newblock In {\em ICCV}, pages 945--953, 2015.

\bibitem{hancCraft2}
J.~Sun, M.~Ovsjanikov, and L.~Guibas.
\newblock A concise and provably informative multi-scale signature based on
  heat diffusion.
\newblock In {\em Computer graphics forum}, volume~28, pages 1383--1392. Wiley
  Online Library, 2009.

\bibitem{volum3}
M.~Tatarchenko, A.~Dosovitskiy, and T.~Brox.
\newblock Octree generating networks: Efficient convolutional architectures for
  high-resolution 3d outputs.
\newblock In {\em ICCV}, pages 2088--2096, 2017.

\bibitem{DGCNN}
Y.~Wang, Y.~Sun, Z.~Liu, S.~E. Sarma, M.~M. Bronstein, and J.~M. Solomon.
\newblock Dynamic graph cnn for learning on point clouds.
\newblock {\em arXiv preprint arXiv:1801.07829}, 2018.

\bibitem{3DGAN}
J.~Wu, C.~Zhang, T.~Xue, B.~Freeman, and J.~Tenenbaum.
\newblock Learning a probabilistic latent space of object shapes via 3d
  generative-adversarial modeling.
\newblock In {\em NIPS}, pages 82--90, 2016.

\bibitem{modelnet}
Z.~Wu, S.~Song, A.~Khosla, F.~Yu, L.~Zhang, X.~Tang, and J.~Xiao.
\newblock 3d shapenets: A deep representation for volumetric shapes.
\newblock In {\em CVPR}, pages 1912--1920, 2015.

\bibitem{FoldingNet}
Y.~Yang, C.~Feng, Y.~Shen, and D.~Tian.
\newblock Foldingnet: Point cloud auto-encoder via deep grid deformation.
\newblock In {\em CVPR}, pages 206--215, 2018.

\end{thebibliography}

\end{document}